\documentclass[preprint,12pt]{elsarticle}

\usepackage[utf8]{inputenc}
\usepackage[T1]{fontenc}
\usepackage[margin=1.25in]{geometry}
\usepackage{graphics} 
\usepackage{epsfig} 
\usepackage{amsmath} 
\usepackage{amsfonts}
\usepackage{hyperref}
\usepackage{subcaption}
\usepackage[numbers]{natbib}

\journal{Robotics and Autonomous Systems}

\begin{document}
\begin{frontmatter}

\title{Evolving Drone Morphologies for Agile Flight}

\author{Elijah Hao Wei Ang}
\ead{e.h.w.ang@tudelft.nl}

\author{Robin Ferede}
\ead{r.ferede@tudelft.nl}

\author{Christophe De Wagter}
\ead{c.dewagter@tudelft.nl}

\author{Guido C. H. E. de Croon}
\ead{g.c.h.e.deCroon@tudelft.nl}

\affiliation{organization={Faculty of Aerospace Engineering, Delft University of Technology},
addressline={Kluyverweg 1},
postcode={2629 HS},
city={Delft},
country={The Netherlands}}

\begin{abstract}
The design of multicopter drones has remained almost the same since their inception. While conventional designs work well in many cases, they may not be optimal in specific tasks. This paper revisits rotary drone design by exploring which body morphologies are best suited for different objectives and constraints. An evolutionary algorithm is used to produce drone morphologies that performs better than standard configurations at thrust-to-weight ratio, maneuverability and size. To generate a range of drones with performance trade-offs between them, the non-dominated sorting genetic algorithm II is used. The evolved drone bodies outperform a standard multicopters in at least one of the three objectives. The Pareto front extrema show improvement of 489.2\%, 23\% and 4.8\% in maneuverability, thrust-to-weight ratio and size, respectively, when compared to a standard quadcopter. The improvement in maneuverability is attributed to the tilt angles of the motors, while the increase in thrust-to-weight ratio is due to the higher number of propellers. Experiments show that the performance improvement transfers to a real world drone racing task, where the evolved drone outperforms the standard drone in terms of number of gates passed, flight speed and lap times.
\end{abstract}


\newpageafter{abstract}

\end{frontmatter}

\section{Introduction}
In recent years, multicopter drones have been widely adopted for various applications, such as surveillance, site inspection and disaster response \cite{hassanalian2017classifications}. Since their inception, many designs have been proposed, but now most drones have adopted the same simple and conventional designs. An example of this is the standard quadcopter. The advantages of these designs are that they have been researched and developed extensively, they are simple to construct, and the implementation of controls is intuitive \cite{bbvl2016survey, paredes2021development}. While such drones may perform adequately for a wide range of applications, they may not be the most optimal design for specific use cases.

Nature and biology have proven that the morphology varies wildly across species \cite{grant1976darwin}. The morphology of each species has been adapted and optimized to produce the best performance needed for its specific environmental niche. This optimization is performed via the natural selection process, where the best-performing individuals are given the opportunity to reproduce and pass on their genes to the next generation, while the worst-performing individuals get eliminated \cite{forrest1993genetic}. This process occurs over multiple generations, resulting in morphologies that are highly adapted to their specific environmental niche.

\begin{figure}
\centering
\includegraphics[width=0.80\textwidth]{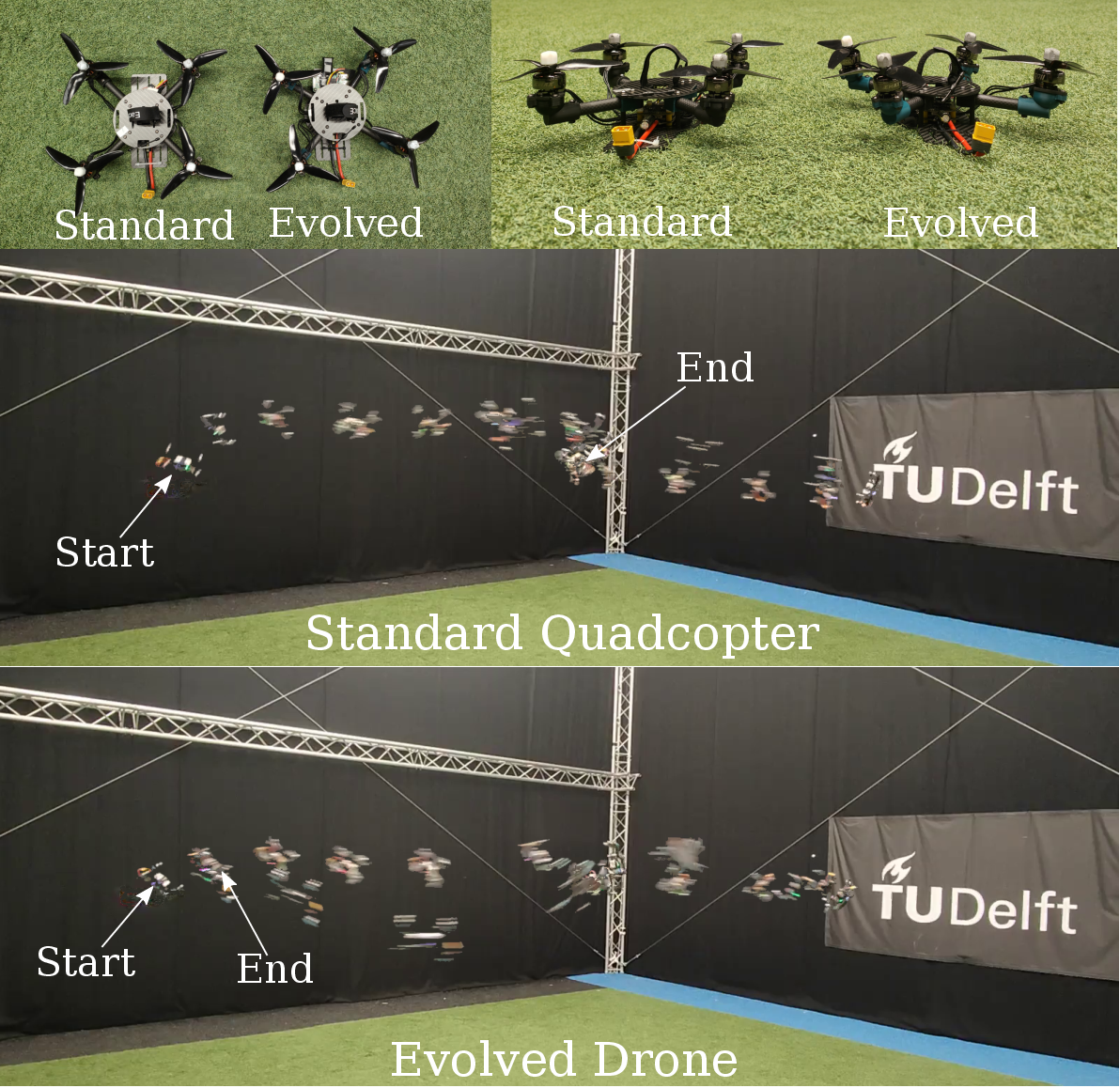}
\caption{We use a multi-objective evolution algorithm to generate novel propeller-based drone designs. In this study, the most maneuverable drone produced by the evolution process is selected and its performance is compared to that of a standard quadcopter in a drone racing task (top). The top-left and top-right images illustrate the differences between the standard and the evolved drone, featuring different motor positions and propeller orientations for the latter. The bottom images show the flight paths of the standard quadcopter (middle) and the evolved drone (bottom) over the same duration. In the same time that the standard quadcopter only manages to complete part of a lap, the evolved drone successfully completes a full lap.}
\label{fig:intro}
\end{figure}

Drawing inspiration from nature, the designs of various systems can also be optimized via the same evolutionary process. This method has proven to be successful in various systems such as antenna design \cite{hornby2011computer}, walking robots \cite{vujovic2017evolutionary}, modular robots \cite{kulz2024optimizing} and recently on morphing wing drones \cite{bergonti2024co}.

In this paper, we explore whether an artificial evolution process can improve upon the design of the most occurring drone type: the multicopter. Specifically, a multi-objective approach to evolve drone morphology is presented. The non-dominated sorting genetic algorithm II (NSGA-II) is used for the evolution of drone bodies. To ensure that the drones produced from the evolutionary algorithm are physically viable, their fitness will be evaluated using model-based objective functions, such as hovering capabilities, maneuverability, thrust-to-weight ratio and size. Thereafter, to demonstrate that drones evolved using these metrics translate to real-world performance improvements, we select the most maneuverable drone to demonstrate fast and agile flight around a drone racing track using end-to-end reinforcement learning policies.

The main contributions of this paper are (1) an extremely computationally efficient model-based fitness evaluation for the evolution of drones, (2) the evolution of drone morphology with guaranteed viability, (3) multi-objective evolution which produces a diverse range of Pareto optimal drones and (4) reinforcement learning of evolved drones to achieve agile flight in real-world tests. The rest of the paper is organized as follows. Section \ref{sec:related_work} looks into related work found in the literature. Section \ref{sec:objective_functions} presents model-based objective functions for drones with arbitrary configurations. Section \ref{sec:evolution_drone} describes the evolution process as well as the genotype to phenotype mapping. Section \ref{sec:evolution_result} discusses the results obtained from evolving drone bodies. Next, sections \ref{sec:experimental_validation} and \ref{sec:experimental_setup} describe the reinforcement learning and experimental setup used for the real-world tests, respectively. Section \ref{sec:experimental_results} presents and discusses the results obtained from the experiments. Lastly, Section \ref{sec:conclusion} concludes this paper.

\section{Related Work}
\label{sec:related_work}
\subsection{Evolutionary Robotics}
The manual design and optimization of systems is often expensive, time consuming and requires significant domain knowledge. As an alternative, several researchers have drawn inspiration from nature by adopting evolutionary processes to optimize their systems. Hornby et al. \cite{hornby2011computer} proposed a novel antenna designed using an evolutionary algorithm that performs better than human-designed antennas. Vujovic et al. \cite{vujovic2017evolutionary} combined evolution with development stages for the evolution of real-world walking robots. The development stage simulates the growing process from infancy to adulthood for each species.

When it comes to drones, evolutionary algorithms have been successful in optimizing the controller, or the "brain", of the drones. For instance, Shamshirgaran et al. \cite{shamshirgaran2021evolutionary} performed multi-objective evolution using particle swarm optimization (PSO) and biogeography-based
optimization (BBO) to tune the parameters of a PD controller for a quadcopter. Similarly, Yazid et al. \cite{yazid2019position} used evolutionary algorithms such as genetic algorithm (GA), PSO and artificial bee colony (ABC) to tune a Takagi-Sugeno-Kang fuzzy logic controller for position control of a quadcopter. Ang and Ng \cite{ang2024genetic}
used a genetic algorithm to separately optimize a linear quadratic Gaussian (LQG) controller and a neural network controller to correct gust-induced deflections and displacements on a flexible wing unmanned aerial vehicle. Additionally, the optimization of drone bodies has been investigated in the literature.  Du et al. \cite{du2016computational} produced an optimized multicopter from an initial user design using convex optimization algorithms. Their algorithm took user-provided designs such as the quadcopter and pentacopter and modified the geometry for increased flight time and payload, respectively. Similarly, Carlone and Pinciroli \cite{carlone2019robot} co-designed drone bodies and hardware using linear programming methods. Their algorithm selects the best components, such as actuators and sensors, to meet specific objectives and constraints. On the other hand, Bergonti et al. \cite{bergonti2024co} optimized the topology and control of morphing wing drones using an evolutionary algorithm. Their simulation results showed that the evolved morphing wing drone outperforms a standard fixed-wing aircraft in terms of energy efficiency and mission time. 

While optimization of drone bodies has been investigated in the past, the most common methods involve linear and non-linear programming, while evolutionary algorithms are seldom explored. Additionally, much of the past research optimized a single objective function at a time, which restricted the diversity of drone designs and the insight into trade-offs in the design space. In our work, we provide an evolution-based morphology optimization of multicopters with multiple degrees of freedom. We also consider multiple objectives, with the aim of generating a diverse range of drone designs that excel in various scenarios while achieving more significant advances over the common drone design. Furthermore, we validate our approach by constructing and deploying the evolved drone morphology in real-world tests.

\subsection{Autonomous Agile Flight}
Conventionally, autonomous agile drone flights are achieved using a combination of trajectory generation \cite{hehn2015real, zhou2023efficient, mellinger2011minimum, tal2020accurate} with optimal control for trajectory tracking, using methods such as nonlinear model predictive control (NMPC). For example, Nguyen et al. \cite{nguyen2024tinympc} implemented model predictive control on the Crazyflie platform to achieve agile flight in a figure-eight trajectory tracking task, while Faessler et al.\cite{faessler2017differential} used a differential flatness based controller for high-speed trajectory tracking on a racing drone.

However, a key limitation of using optimal control methods for agile trajectory tracking is the need to separate the problem into multiple layers, which include the reference trajectory generation (planning) and tracking (control). Furthermore, the control problem is often split into high-level control to compute the desired accelerations and rates to minimize tracking errors, and low-level control to produce actuation commands. While each layer may be optimized individually, their combination often lead to sub-optimal performance. To overcome this limitation, Izzo et al. \cite{izzo2024optimality} introduced a guidance and control network (G\&CNet) which is trained to directly map the system's state to actions. Applying it to drones, Ferede et al. \cite{ferede2024net} trained an adaptive G\&CNet that maps the quadcopter's states directly to motor commands via supervised learning. A dataset of state-action pairs is generated by solving an energy minimization problem constrained by the quadcopter's dynamics and actuation limits. 

In recent years, deep reinforcement learning has been employed to achieve autonomous agile flight. For example, Song et al. \cite{song2021autonomous} achieved high-speed autonomous flight through a variety of drone racing tracks by training their models using Proximal Policy Optimization (PPO). Similarly, Mengozzi et al. \cite{mengozzi2024towards} trained a reinforcement learning policy for agile flight in dynamic environments. When compared to model predictive control (MPC) based method, the learned policy achieved an increase in success rate by 95\%. Research has shown that for agile flight scenarios such as in drone racing, reinforcement learning outperforms optimal control methods by achieving higher success rates and better lap times in real world implementation \cite{song2023reaching}. Notably, Kaufmann et al. \cite{kaufmann2023champion} developed an autonomous drone racing system that achieved performance comparable to professional human pilots. This was achieved by combining a perception module trained to estimate the drone's state from visual and sensor data, and a control policy trained using deep reinforcement learning, which maps the estimated states to actions. Their autonomous drone won several races against champion pilots, and also achieved the fastest race time in their experiment.

While autonomous agile flight has been successfully demonstrated in prior work, the methods were primarily focused on standard drone designs, such as the quadcopter. Restricting the morphology to conventional designs limits the potential for further performance improvements. In this paper, we aim to go beyond these limits by optimizing the physical design of the drones.

\section{Model-Based Objective Functions}
\label{sec:objective_functions}
Although drone mission objectives can be extremely diverse and complex, ultimately a drone's capabilities are very much related to low-level physical design properties. For instance, mission capabilities such as flight time or range are related to thrust-to-weight ratio, while agility is related to the moments that can be generated by the propellers. In our study, we define objectives in terms of such low-level physical design properties. The additional advantage of this approach is that it enables the extremely efficient computation of the objectives, resulting in quick evolution.

In this section, we explain how we model the drones in order to calculate design performance objectives.

\subsection{Actuator Effectiveness Matrices}
The static thrust and torque produced by a rotor and propeller are directly proportional to the square of the rotational speed of the propeller, and are given as,
\begin{equation}
    \begin{split}
        T = k_f \omega^2 \\
        M = k_m \omega^2
    \end{split}
\end{equation}
where $\omega$ is the rotation speed of the propeller, and $k_f$ and $k_m$ are the thrust and torque coefficients of the propeller, respectively. It is convenient to define an actuator command $\eta \in [0,1]$ which maps linearly to $[0,\omega^2_{max}]$ in order to linearize the equations. The forces and moments acting on the drone due to motor $i$ is given as,

\begin{equation}
    \begin{split}
        \mathbf{f}_i &=\frac{T_i}{m}\mathbf{n}_i \\
        \mathbf{m}_i &= \mathbf{J}^{-1}(\mathbf{r}_i \times T_i\mathbf{n_i} - M_i R_i \mathbf{n}_i)
    \end{split}
\end{equation}
where $m \in \mathbb{R}$ and $\mathbf{J} \in \mathbb{R}^{3 \times 3}$ are the mass and moment-of-inertia of the entire drone, respectively. $T_i$ and $M_i$ are the thrust and torque produced by motor $i$, respectively. $\mathbf{n}_i \in \mathbb{R}^3$ is the motor orientation of motor $i$ with respect to the body reference frame fixed to the center of gravity, $\mathbf{r}_i \in \mathbb{R}^3$ is the position vector of the motor in body reference frame, and $R_i \in \{-1, +1\}$ indicates the propeller rotation direction ($+1$ when rotation axis defined using the right-hand rule is in the same direction as thrust). By combining the effects of all propellers, the forces and moments acting about the center of gravity of the drone can be condensed into a single matrix equation. For a drone with $n$ propellers, the specific forces and moments are defined as,

\begin{equation}
    \begin{split}
        \mathbf{f} = \mathbf{B}_f \boldsymbol{\eta} \\
        \mathbf{m} = \mathbf{B}_m \boldsymbol{\eta} 
    \end{split}
    \label{eqn:forces}
\end{equation}
where $\boldsymbol{\eta} \in \mathbb{R}^n$ is the vector of actuator commands, $\mathbf{f} \in \mathbb{R}^3$ and $\mathbf{m} \in \mathbb{R}^3$ are the specific force and moment vectors, respectively, and $\mathbf{B}_f \in \mathbb{R}^{3 \times n}$ and $\mathbf{B}_m \in \mathbb{R}^{3 \times n}$ are the force and moment actuator effectiveness matrices, respectively.

\subsection{Static Hovering}
In this work, a drone is deemed feasible only if it satisfies the minimum condition of being able to hover. In order to hover, the drones must produce sufficient thrust to overcome their weight and maintain their attitude. Allowing drones with arbitrary configurations may result in multiple hovering attitudes. Hence, the aim is to find an optimized set of control inputs $\hat{\boldsymbol{\eta}}$ from control space $\boldsymbol{\eta} \in \mathbb{R}^n$ which allows the drone to hover statically, i.e. without rotating, at minimum cost. If a solution cannot be found, the drone is unable to hover statically. This can be defined as a non-linear programming problem as follows,
\begin{equation}
    \begin{split}
        \text{min} \quad &J = \hat{\boldsymbol{\eta}}^T \hat{\boldsymbol{\eta}} \\
        \text{s.t.} \quad &||\mathbf{B}_f \hat{\boldsymbol{\eta}}||_2 = g \\
        &\mathbf{B}_m \hat{\boldsymbol{\eta}} = \mathbf{0} \\
        &\underline{\boldsymbol{\eta}} \leq \hat{\boldsymbol{\eta}} \leq \overline{\boldsymbol{\eta}}. \\
    \end{split}
\end{equation}
where $g$ is the gravitational acceleration, $\underline{\boldsymbol{\eta}}$ is the lower actuation limit and $\overline{\boldsymbol{\eta}}$ is the upper actuation limit. The first constraint specifies that the optimal control input has to produce enough thrust to overcome gravity in order to hover. The second constraint states that the optimal control input has to result in zero resultant moments acting on the body of the drone while hovering. The final equation ensures that the control inputs are kept between the upper and lower actuation limits. The sequential least-squares programming (SLSQP) algorithm \cite{kraft1988software} is used to find the optimal control inputs.

\subsection{Non-static Hovering}
In some configurations, the drones may not be able to hover statically without rotating. Instead, these drones can maintain their position while spinning about their yaw axis. A clear example of this is a tricopter without a tilting tail rotor. While it is not optimal, such designs should not be discarded as they may still contain desirable properties, such as the ability to hold their altitude, which may be carried over to the next generation of drones during evolution. Hence, a set of control inputs $\hat{\boldsymbol{\eta}}$ from control space $\boldsymbol{\eta} \in \mathbb{R}^n$ which allows the drone to hover while spinning at minimum cost can be computed using the following constraints,
\begin{equation}
    \begin{split}
        \text{min} \quad &J = \hat{\boldsymbol{\eta}}^T \hat{\boldsymbol{\eta}} \\
        \text{s.t.} \quad &||\mathbf{B}_f \hat{\boldsymbol{\eta}}||_2 = g \\
        &||\mathbf{B}_f \hat{\boldsymbol{\eta}}  \times \mathbf{B}_m \hat{\boldsymbol{\eta}}||_2 = 0 \\
        &\underline{\boldsymbol{\eta}} \leq \hat{\boldsymbol{\eta}} \leq \overline{\boldsymbol{\eta}} \\
    \end{split}
\end{equation}
The first and the last constraints are the same as for static hovering. The second constraint no longer requires zero resultant moment acting on the body. Instead, it requires that the direction of thrust and the axis of rotation are parallel, resulting in a drone that spins while hovering. This is done by ensuring that the magnitude of the cross product between the thrust and moment vectors has a magnitude of zero.

\subsection{Thrust-to-weight Ratio}
The thrust-to-weight ratio is used as a proxy for several goals at once.
With a higher thrust-to-weight ratio, the drone can either fly faster or carry larger batteries for more flight time, carry more payload for better mission results or carry better sensors and processing. These all lead to improved mission performance which forms a secondary optimization goal.
Once $\hat{\boldsymbol{\eta}}$ has been computed, the maximum possible thrust-to-weight ratio while maintaining attitude and heading can be computed by scaling the control inputs until control input saturation is reached for at least one of the propellers. Since the control inputs are scaled to a range of $[0,1]$, this can be computed using,

\begin{equation}
    \hat{\boldsymbol{\eta}}_{max} = \frac{\hat{\boldsymbol{\eta}}}{\displaystyle\max_{i} \hat{\eta_i}}
\end{equation}
where $\hat{\boldsymbol{\eta}}_{max}$ gives the control input which results in largest thrust-to-weight ratio without exceeding the upper or lower bounds and $\hat{\mathbf{\eta}}_i$ gives the $i$th control input. Hence, the thrust-to-weight ratio is given as,

\begin{equation}
    \alpha = \frac{||\mathbf{B}_f \hat{\boldsymbol{\eta}}_{max}||_2}{g}
\end{equation}
where $\alpha$ is the thrust-to-weight ratio.

\subsection{Controllability and Maneuverability}
The controllability determines if a drone has full moment actuation. Typically, a drone is controllable if the controllability matrix has full row rank. However, this measure does not provide a measure of the agility or maneuverability of the drone. Instead, a better indicator can be obtained by using the controllability Gramian. For the dynamics given by Eq. \ref{eqn:forces}, the Gramian is computed as,

\begin{equation}
    \mathbf{W}_c = \mathbf{B}_m \mathbf{B}_m^T 
\end{equation}
where $\mathbf{W}_c \in \mathbb{R}^{3 \times 3}$ gives the controllability Gramian \cite{brunton2022data}. The eigenvalues of the Gramian give an indication on the maneuverability of each of the principal axes defined by their corresponding eigenvectors. With higher maneuverability, drones will be able to perform aggressive maneuvers, which is useful in applications such as drone racing or physically interacting with the world. As a more agile drone is desired, E-optimality is used for this objective, where the maneuverability of the drone will be characterized by the smallest eigenvalue, which corresponds to its least maneuverable principal axis, given as,

\begin{equation}
    \lambda = \text{min} \ \text{eig}(\mathbf{W}_c)
\end{equation}
where larger $\lambda$ values correspond to more maneuverable drones.

\subsection{Size of Drone}
Generally, smaller drones are preferred as they will be able to navigate and squeeze through small spaces. However, the trade-off is that a smaller drone may not have the space to accommodate heavier equipment such as high-resolution cameras or GPU-enabled companion computers. The size of a drone can be parameterized using the volume of its convex hull formed by the propeller positions. In 2D cases where propellers are placed on the same plane, the area of the convex hull will be used instead.

\section{Evolution of Drone Bodies}
\label{sec:evolution_drone}
\subsection{Phenotype and Genotype}
The phenotype of the drone defines the location and direction of the propellers. In this study, the parameters of the drones to be evolved are given as follows:
\begin{enumerate}
\item number of propellers: $n \in \{4,5,6,7,8\}$
\item arm length: $l \in [0.1 \, \text{m}, 0.3 \, \text{m}]$
\item arm angle: $\theta \in [-180^{\circ}, 180^{\circ}]$
\item propeller inclination: $\varphi \in [0^{\circ}, 15^{\circ}]$
\item propeller azimuth wrt. arm: $\psi \in [-90^{\circ}, 90^{\circ}]$
\item propeller direction: $R \in \{\text{'CCW'}, \text{'CW'}\}$
\end{enumerate}
In this study, for manufacturability purposes, we use the constraint that all motors are placed on the $x-y$ plane of the drone. The minimum number of propellers is 4 as that is the minimal condition required for a drone to achieve basic actuation \cite{hamandi2021design}. 5-inch propellers are used in all cases. The physical parameters used in the evolution process are obtained from manufacturer data and are presented in Table \ref{tab:evolution_params}

\begin{table}[t]
    \centering
    \caption{Physical parameters used for evolution}
    \begin{tabular}{c|c}
        \hline
        \textbf{Parameter} & \textbf{Value} \\
        \hline
        Main body weight & 300 g \\
        Arm mass per unit length & 75 g/m \\
        Propeller and motor mass & 33.5 g \\
        Propeller thrust coefficient & 1.08 $\times$ 10\textsuperscript{-6} \\
        Propeller torque coefficient & 1.22 $\times$ 10\textsuperscript{-8} \\
        \hline
    \end{tabular}
    
    \label{tab:evolution_params}
\end{table}

The genotype is a direct representation of the phenotype and can be broken down into eight subsections, one for each propeller. Each value in the subsection is within the range of $[-1, 1]$, and is directly mapped to the parameter intervals defined above. Continuous parameters (parameters 2 to 5) are linearly mapped, while discrete parameters (parameters 1 and 6) are first discretized evenly before mapping. To keep the genotype length the same while varying the number of propellers, only the first $n$ subsections are mapped into propellers in the phenotype. The remaining subsections stay in the genotype and act as "phantom" propellers which have a chance of being activated during the evolution process. In the event that propellers intersect each other, all arm lengths will scale up until sufficient distance between propellers is achieved. It is important to note that this scaling is part of the genotype-to-phenotype mapping process, and the genotype remains the same after scaling.

\subsection{NSGA-II and Fitness Evaluation}
The initial population of genotypes is generated using the Latin hypercube sampling method to ensure distributed sampling across the search space. The evolution of the drones is performed using the NSGA-II algorithm, which is an elitist multi-objective evolutionary algorithm that ranks individuals based on non-dominated sorting \cite{deb2002fast}. 

As NSGA-II is a multi-objective optimization algorithm, the fitness function is not determined by a single metric. Instead, multiple objective functions are computed and individuals are ranked by the number of individuals that dominate them. Hence, individuals who get dominated the least will be ranked the highest, producing a range of Pareto optimal results. Subsequent rankings will result in their own Pareto front. NSGA-II is used as it provides a fast algorithm for the non-dominated sorting as well as solution density estimation to preserve diversity within the population. The three objective functions in this experiment are to maximize (1) thrust-to-weight ratio and (2) maneuverability while minimizing (3) the volume of the convex hull of the drone.  Additionally, drones with the ability to hover statically are ranked ahead of drones that can only achieve spinning hover, which are, in turn, ranked ahead of drones that cannot hover. The code and implementation of the evolution process are available on GitHub\footnote{\url{https://github.com/tudelft/nsga-drone-evolve}}.

\section{Evolution Results and Discussions}
\label{sec:evolution_result}
An initial population of size 600 is evolved over 2000 generations using NSGA-II. A value-wise arithmetic crossover scheme with a 20\% mutation rate is used to populate the next generation in the evolutionary process. Parents are selected using the binary tournament selection algorithm. Simulation results obtained from the evolution process are compared with a conventional 5-inch drones with four to eight propellers.

\subsection{Evolution History}
\begin{figure}
\centering
\includegraphics[width=0.70\textwidth]{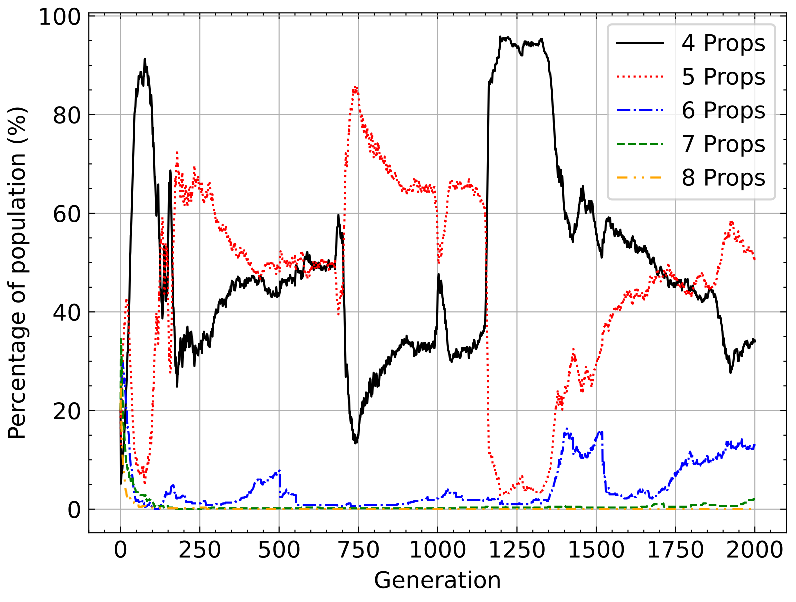}
\caption{Percentage of number of arms for each generation.}
\label{fig:fullsampledarms}
\end{figure}

Figure \ref{fig:fullsampledarms} shows the proportion of drones with varying numbers of propellers across the generations. The initial random population starts with approximately 20\% of each propeller number. However, as the evolution progresses, the number of 8-propeller drones quickly diminishes. As a result, the remaining drones have between 4 to 7 propellers, with the 5-propeller drones making up more than half of the population. It is observed that for the majority of the generations after 150, drones with 7 propellers are fully eliminated. However, due to the mutation process, 7-propeller drones start to reappear and survive the evolution process, and make up 2\% of the population. On the other hand, 8-propeller drones are unable to benefit from mutation.

\subsection{Pareto Optimal Drones}
\begin{figure}
\centering
\includegraphics[width=0.60\textwidth]{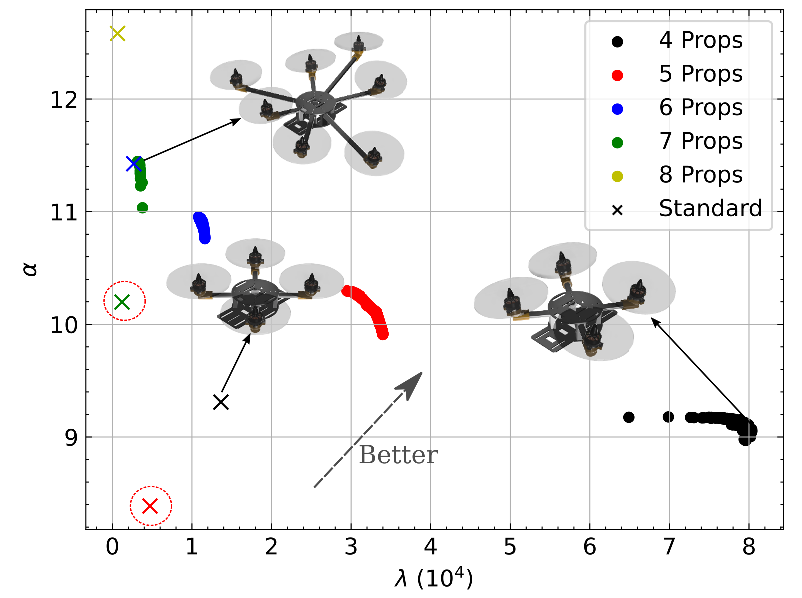}
\caption{Scatter plot of Pareto optimal and standard drones' thrust-to-weight ratio ($\alpha$) against maneuverability ($\lambda$) at generation 2000. Standard drones marked in dotted red circles are dominated.}
\label{fig:fullsampled-ctrl-alpha}
\end{figure}

\begin{figure}
\centering
\includegraphics[width=0.60\textwidth]{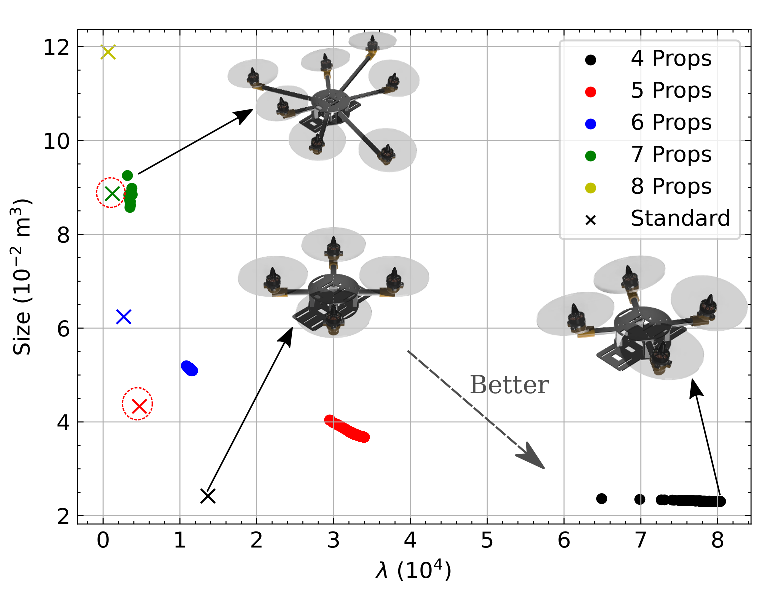}
\caption{Scatter plot of Pareto optimal and standard drones' size against maneuverability at generation 2000. Standard drones marked in dotted red circles are dominated.}
\label{fig:fullsampled-ctrl-size}
\end{figure}

\begin{figure}
\centering
\includegraphics[width=0.60\textwidth]{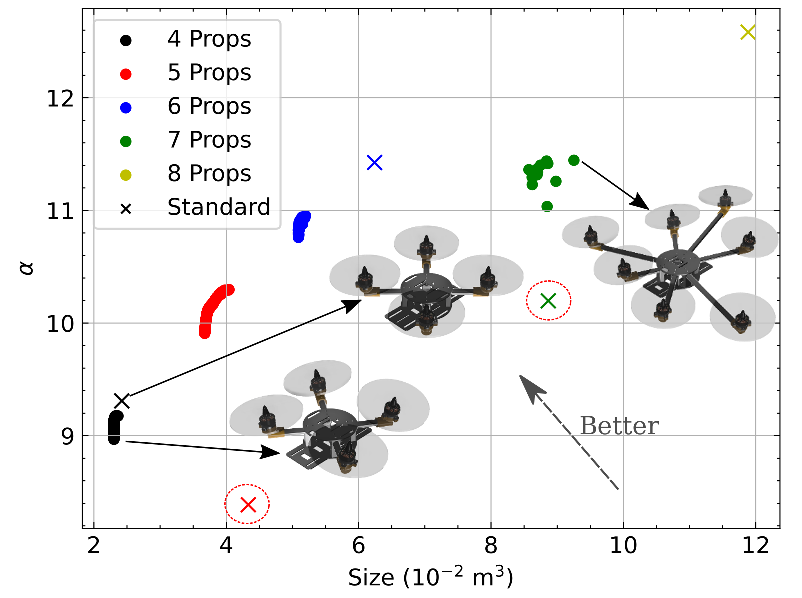}
\caption{Scatter plot of Pareto optimal standard drones' thrust-to-weight ratio ($\alpha$) against size at generation 2000. Standard drones marked in dotted red circles are dominated.}
\label{fig:fullsampled-size-alpha}
\end{figure}

\begin{figure}
    \centering
    \begin{subfigure}[b]{0.60\textwidth}
        \centering
        \includegraphics[width=\textwidth]{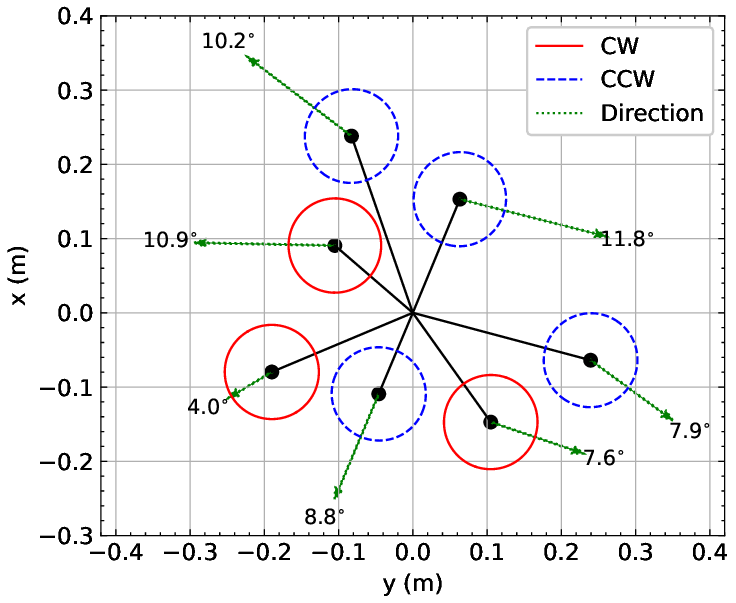}
        \caption{Highest thrust-to-weight ratio drone.}
        \label{fig:alpha-drone}
    \end{subfigure}
    \hfill
    \begin{subfigure}[b]{0.60\textwidth}
        \centering
        \includegraphics[width=\textwidth]{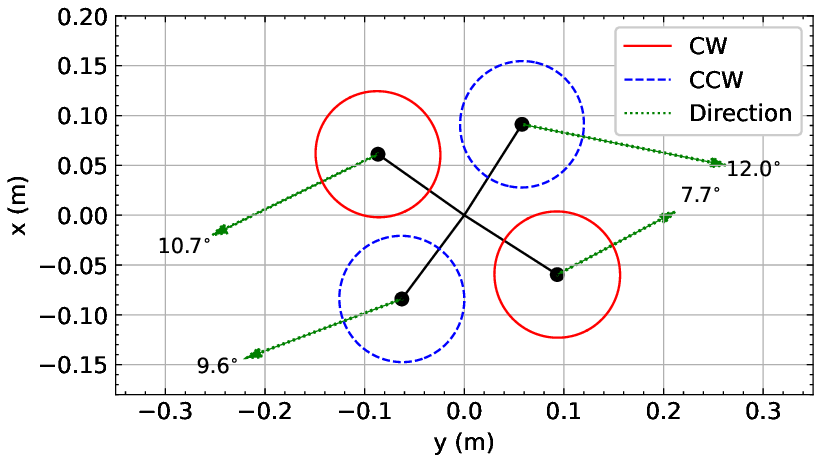}
        \caption{Smallest and most maneuverable drone.}
        \label{fig:ctrl-drone}
    \end{subfigure}
    \caption{Pareto extrema drones after 2000 generations.}
    \label{fig:pareto-extrema}
\end{figure}

Figures \ref{fig:fullsampled-ctrl-alpha}, \ref{fig:fullsampled-ctrl-size} and \ref{fig:fullsampled-size-alpha} show the Pareto front at the end of evolution projected on the $\lambda - \alpha$, $\lambda - \text{size}$ and $\text{size} - \alpha$ planes, respectively. Looking at the plots, size, maneuverability and thrust-to-weight ratio are very clearly clustered according to the number of propellers, making it a key parameter determining the drone's characteristics. Drawing comparisons to nature, each cluster acts as a distinct species, where each individual within a species tends to have similar characteristics. Figures \ref{fig:alpha-drone} and \ref{fig:ctrl-drone} show the drone with the highest thrust-to-weight ratio and the smallest drone with the highest maneuverability, respectively.

Generally, increasing the number of propellers results in a larger drone with higher thrust-to-weight ratio, but reduced maneuverability. This highlights the presence of conflicting objectives, where drones with increased thrust-to-weight ratios tend to be larger and less maneuverable. On the other hand, decreasing the size leads to an increase in maneuverability, which are both desirable according to the objectives. This shows that given the current phenotype space, it is impossible to simultaneously maximize all three objectives, presenting a trade-off between size, thrust-to-weight ratio and maneuverability for the evolved population.

Comparing the results with the standard quadcopter, the evolution process managed to obtain drones that outperform the quadcopter in each of the three objectives. The drone with the highest $\alpha$ presents an improvement of 23\% to the thrust-to-weight ratio while the most maneuverable drone shows a significant increase in maneuverability by 489.2\%. The increase in thrust-to-weight ratio is largely due to the use of additional propellers, while the increase in maneuverability can be attributed to tilted propellers and smaller drones for increased specific moments, especially about the yaw axis. The reduction in size for the smallest drone is 4.8\%, which is small compared to the improvement in thrust-to-weight ratio and maneuverability. Drones with high maneuverability will be beneficial in situations where fast and tight maneuvers are required, such as in drone racing. On the other hand, having a high thrust-to-weight ratio is necessary for performing tasks such as navigation or simultaneous localization and mapping (SLAM) with heavy equipment such as depth cameras and powerful onboard computers. Lastly, small drones are generally preferred when needed to navigate through dense and cluttered environments such as unmanaged forests. 

While the evolved drones manage to dominate the quadcopter in Figures \ref{fig:fullsampled-ctrl-alpha} and \ref{fig:fullsampled-ctrl-size}, the evolution process is unable to obtain drones that are smaller yet possess a higher thrust-to-weight ratio than a standard quadcopter. This is primarily because tilting the motors in different directions reduces the effective vertical thrust. As a result, the only way to increase thrust-to-weight ratio further is by adding more propellers, which inherently increases the size of the drone. In this regard, the standard quadcopter, hexacopter and octocopter remain as non-dominated designs. 

\subsection{Discussion}
As seen from the results in Figure \ref{fig:fullsampledarms}, drones with 8 propellers are eliminated by the end of the evolution process, despite having equal proportions of 4 to 8-propeller drones at the beginning. Larger drones generally result in lower maneuverability, as shown in Figure \ref{fig:fullsampled-ctrl-size}. The angular acceleration due to a single propeller can be modeled as,
\begin{equation}
    \dot{\Omega} \propto \frac{l}{ml^2 + \frac{\mu}{3} l^3 + I_{fc}}
    \label{eqn:angularaccel}
\end{equation}
where $\dot{\Omega}$ gives the angular acceleration, $l$ gives the length of the arm, $m$ is the mass of the motor for the propeller, $\mu$ is the mass per unit length of the arm connecting the motor to the body, and $I_{fc}$ is the moment of inertia due to the battery and flight controller stack. The numerator represents the moments produced while the denominator represents the moment of inertia of the system. While the moment generated by the propeller increases linearly with the length of the arm, the resulting moment of inertia increases quadratically with the length, which results in an overall reduction in the angular acceleration. Moreover, propellers placed further apart require longer arms which further add to the mass and moment of inertia of the drone, which is represented by a cubic relationship. The angular acceleration given by Eq. \ref{eqn:angularaccel} increases to a maximum value within a very narrow band of $l$ which lies outside of the phenotype mapping range before decreasing with further increase of the arm length. As a result, the most maneuverable drone is also the smallest drone after evolution, as shown in Figure \ref{fig:ctrl-drone}.

On the other hand, larger drones tend to have a higher thrust-to-weight ratio. Generally, having more propellers will result in a higher thrust-to-weight ratio, which also requires the drones to be larger to prevent propellers from intersecting each other. Despite this, only a few 7 or 8-propeller drones emerged from the evolution process. Apart from having a high number of propellers, the propellers need to be evenly spaced apart and evenly actuated to achieve a high thrust-to-weight ratio. In cases where the propellers are not evenly spaced apart, some propellers may require higher actuation to counter the moments from other propellers while hovering, which will result in earlier saturation of these propellers. In such cases, the maximum possible thrust-to-weight ratio is reduced.

The evolutionary process is unable to obtain the drone with the theoretically highest thrust-to-weight ratio, which is an octocopter with all propellers pointing vertically up, spaced evenly apart. This is because the viable and optimal solution space for 8-propeller drones is very small as compared to drones with fewer propellers. This results in difficulty in the search for optimal drones with 8 propellers despite having a high mutation rate of 20\% for exploration, resulting in them getting eliminated from the population very early on.

While the evolutionary process is effective in obtaining drones that outperform a standard quadcopter, hexacopter and octocopter in at least one objective, it was unable to fully dominate the quadcopter. The design of the standard drones has arms positioned at equal distances from each other, which allows for shorter arms to be used without the propellers impeding each other. At the same time, the propellers are all pointing vertically upwards. The combination of these two results in a sufficiently small drone with the highest thrust-to-weight ratio possible for a 4, 6 and 8-propeller drone. However, this leads to yaw maneuverability being a major weakness of standard drones, as the only source of actuation is from the counter-torque produced from spinning the propellers. 

Interestingly, standard drones with odd number of propellers (5 and 7 propellers) are dominated by the evolved population and do not appear on the Pareto front. With odd number of propellers, there are unequal number of clockwise from counter-clockwise spinning propellers. This results in a fundamental constraint, where it is impossible to achieve zero yaw torque while producing maximum thrust with equal motor speeds. To maintain attitude and heading, the control allocation need to account for the torque imbalance, resulting in unequal actuation and earlier saturation of the motors. Hence, despite having more propellers, the maximum thrust-to-weight ratio achievable is lowered. Combined with its larger size due to additional propellers and reduced maneuverability due to larger moment-of-inertia, standard odd-numbered-propeller drones are dominated across all objectives.

\subsection{Computational Cost}
The computational cost of this method is minimal due to the structure of the objective function evaluation. The objectives are based purely on physical attributes, namely size, maneuverability and thrust-to-weight ratio, as opposed to flight behavior and dynamics. Since the objectives depend solely on the drones' morphology, specifically the positions and orientations of the propellers, they can be efficiently computed using matrix operations on relatively small matrices with maximum dimension of $6 \times 8$. This method is in contrast to simulation-in-the-loop evolution, which requires significantly heavier computation in simulation and policy learning. In this study, evaluating the objectives for a population of 600 individuals requires approximately two seconds per generation, enabling rapid exploration of the morphological space.

\section{Experimental Validation}
\label{sec:experimental_validation}
We investigate whether the low-level objectives translate to better performance on a higher-level tasks. While the objectives of size and thrust-to-weight ratio directly translate to real-world performance and are straight-forward to validate, the maneuverability objective is more abstract and performance differences in real world task are less immediately apparent. To test whether the maneuverability measure translate to actual flight performance, we use reinforcement learning to train neural networks to fly both a standard quadcopter and the evolved drone with highest maneuverability in an agile flight task. Specifically, we evaluate and validate them on tight, twisted lemniscate tracks, which emphasizes the need for large moment authority, described by the maneuverability objective.

\subsection{Multicopter Dynamics}
In order to train the reinforcement learning policy, a multicopter simulator is used, providing a controlled environment for the drone to learn the policies. The equations of motion governing the multicopter dynamics are similar to those in \cite{ferede2024end} and \cite{ferede2025netrulealldomain}. The states and control inputs for the multicopter are defined as
\begin{equation}
    \mathbf{x} = [\mathbf{p}, \mathbf{v}, \boldsymbol{\Theta}, \boldsymbol{\Omega}, \boldsymbol{\omega}]^T \quad \mathbf{u} = [u_1, u_2, \cdots, u_n]^T
\end{equation}
where $\mathbf{p} \in \mathbb{R}^3$ is the position of the drone, $\mathbf{v}\in \mathbb{R}^3$ is the velocity, $\boldsymbol{\Theta}\in \mathbb{R}^3$ are the Euler angles, $\boldsymbol{\Omega}\in \mathbb{R}^3$ give the body rates and $\boldsymbol{\omega}\in \mathbb{R}^n$ are the propeller speeds in rad/s. $\mathbf{u} \in [0,1]^n$ are the normalized motor commands, such that $\eta_i = u_i^2$. The equations of motion of the drone are given as follows:
\begin{equation}
    \begin{split}
        \dot{\mathbf{p}} &= \mathbf{v} \\
        \dot{\mathbf{v}} &= g \mathbf{e_3} + \mathbf{R}(\boldsymbol{\Theta}) \mathbf{F} \\
        \dot{\boldsymbol{\Theta}} &= \mathbf{Q}(\boldsymbol{\Theta}) \boldsymbol{\Omega} \\
        \dot{\boldsymbol{\Omega}} &= \mathbf{M} \\
        \dot{\boldsymbol{\omega}} &= \tau^{-1} (\boldsymbol{\omega_c} - \boldsymbol{\omega})
    \end{split}
    \label{eqn:eom}
\end{equation}
where $\omega_{ci} = (\omega_{max} - \omega_{min}) \sqrt{k_l u_i^2 + (1-k_l)u_i} + \omega_{min}$.

The forces and moments $\mathbf{F}$ and $\mathbf{M}$ are given as follows:
\begin{equation}
    \begin{split}
        \mathbf{F} &= \sum_{i=1}^{n} \left[K_T\left(\omega_i^2 - 2 \pi \frac{\mathbf{v_b} \cdot \mathbf{n}_i}{P}\omega_i\right)\mathbf{n}_i + K_D\mathbf{v}^\bot_i\omega_i\right] \\
        \mathbf{M} &= \sum_{i=1}^{n} [K_{\Omega i}\omega_i^2 + K_{d\Omega i}\dot{\omega}_i]
    \end{split}
\end{equation}
where $K_T = \text{diag}(k_{\omega x}, k_{\omega y}, k_{\omega z})$ is the static thrust coefficient of the propeller, $K_D = \text{diag}(k_{x}, k_{y}, k_{z})$ is the drag coefficient, $K_{\Omega} = \text{diag}(k_{p}, k_{q}, k_{r})$ and $K_{d\Omega} = \text{diag}(k_{dp}, k_{dq}, k_{dr})$ are the moment coefficients. $\mathbf{v_b}$ is the translational velocity in body frame, $\mathbf{n}_i$ is the normal vector for the $i$th propeller in body frame, $P$ is the propeller pitch in m/revolution and $\omega$ is the motor speed. $\mathbf{v}^\bot$ is the projection of the translational velocity along the propeller plane, given as 
\begin{equation}
    \mathbf{v}^\bot = \mathbf{v_b} - (\mathbf{v_b} \cdot \mathbf{n}_i) \mathbf{n}_i.
\end{equation}

The first term represents the thrust given by the propellers adjusted by the normal inflow velocity acting on the propeller, as described by conservation of momentum and simplified blade element theory \cite{ruiz2016model}. The second term gives the drag from the propellers \cite{martin2010true}, also known as rotor drag. 

The values of coefficients in $K_T$, $K_D$, $K_\Omega$ and $K_{d\Omega}$ are specific to individual drone morphologies, and are obtained using system identification of real flight data. This is achieved by performing manual flights that excite all degrees-of-freedom (thrust, roll, pitch, yaw), as well as high speed flights in a circular path to estimate the drag coefficients. As a result, the drone's dynamics are fully parameterized by these coefficients.

\subsection{Problem Formulation}
\begin{figure}
    \centering
    \includegraphics[width=0.35\textwidth]{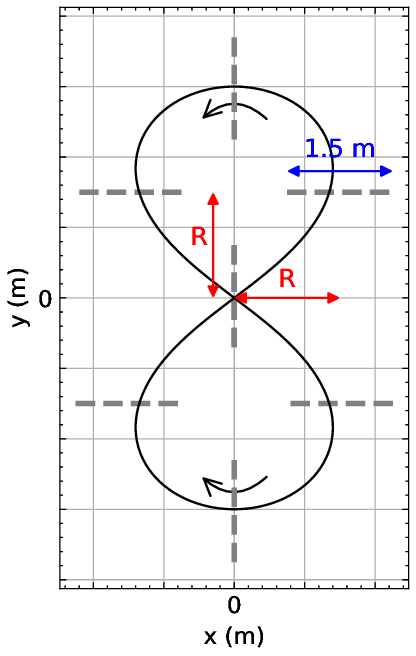}
    \caption{Lemniscate trajectory (black) with seven gates (grey) of size 1.5x1.5m. The size of the track is parameterized by gate distance R.}
    \label{fig:race_track}
\end{figure}

In order to demonstrate improved agility in the real world, both the standard and evolved drones are trained to navigate a drone racing track as quickly as possible. The race track follows a lemniscate trajectory and is defined by seven gates, each measuring 1.5x1.5m, as shown in Figure \ref{fig:race_track}. A lemniscate trajectory is selected for its sharp turns in both directions, which challenges the agility of the drone. Two variations of the tracks are used; a larger track with R=1.5m and a smaller track with R=1.0m. 

The reinforcement learning policy is trained by formulating the problem as a Markov decision process (MDP). The MDP is defined as a tuple $(\mathcal{S}, \mathcal{A}, \mathcal{P}, \mathcal{R}, \gamma)$, where $\mathcal{S}$ represents the state space, $\mathcal{A}$ represents the action space, $\mathcal{P}$ represents the state transition defined by Eq. \ref{eqn:eom}, $\mathcal{R}$ represents the reward function and $\gamma$ represents the discount factor.

The state space $\mathcal{S}$ is the drone's observations, and is given by
\begin{equation}
    \mathbf{x}_{obs} = [\mathbf{p}^{g_i}, \mathbf{v}^{g_i}, \boldsymbol{\Theta}^{g_i}, \boldsymbol{\Omega}, \boldsymbol{\omega}, \mathbf{p}_{g_{i+1}}^{g_i}, \psi^{g_i}_{g_{i+1}}]^T
\end{equation}
where $\mathbf{p}^{g_i}$, $\mathbf{v}^{g_i}$ and $\boldsymbol{\Theta}^{g_i}$ are the position, velocity and orientation of the drone with respect to the current target gate $g_i$, respectively. $\mathbf{p}_{g_{i+1}}^{g_i}$ and $\psi^{g_i}_{g_{i+1}}$ are the position and orientation of the next target gate $g_{i+1}$ with respect to the current target gate, respectively. On the other hand, the action space $\mathcal{A}$ is the normalized motor commands $\mathbf{u}$.

The reward function follows the implementation in \cite{ferede2025netrulealldomain}, which consists of a progress reward, rate penalty and collision penalty. The reward is given as
\begin{equation}
    \mathcal{R}_k = \biggl\{
    \begin{array}{lr}
         -10 & \text{if collided}  \\
         ||\mathbf{p}_{k-1} - \mathbf{p}_{gk}|| - ||\mathbf{p}_{k} - \mathbf{p}_{gk}|| - c||\boldsymbol{\Omega}||& \text{otherwise} 
    \end{array}
    \label{eqn:reward_function}
\end{equation}
where the equation gives the drone a penalty if it collides with the ground or flies out of bounds. When a collision occurs or if the drone misses a gate, the episode is terminated. The first two terms in the second equation give the progress reward, while the last term penalizes excessive body rates. The coefficient for the rate penalty is set as $c=0.001$.

At the start of each episode during training, the drones are placed 1m in front of a randomly selected gate and the state variables are randomly initialized as follows: $\mathbf{v} \in [-0.5,0.5]^3$, $\boldsymbol{\Theta}_{0,1} \in [-\frac{\pi}{9},\frac{\pi}{9}]^2$, $\boldsymbol{\Theta}_2 \in [-\pi,\pi]$, $\boldsymbol{\Omega} \in [-0.1,0.1]^3$ and $\boldsymbol{\omega} \in [\omega_{min}, \omega_{max}]^n$.

\subsection{RL Policy and Training}
In this paper, the PPO algorithm \cite{schulman2017proximal} is selected for the training of the policy. The policy consists of three fully connected layers, each with 64 neurons using the ReLU activation function. The neural network receives the observations $\mathbf{x}_{obs}$ as inputs and output the normalized motor commands $\mathbf{u}$, resulting in an end-to-end control policy.

In order to reduce the sim-to-real gap, we apply domain randomization to the force and moment coefficients during training. This ensures that the policies are robust to model discrepancies and unmodeled effects \cite{tobin2017domain}. In this study, the coefficients vary randomly by $\pm$30\% from the baseline values obtained through system identification. The coefficients are presented in Table \ref{tab:drone_parameters}.

The training is performed in Python by implementing the dynamics given by Eq. \ref{eqn:eom} within a custom Gym environment. The PPO algorithm is implemented using the Python library Stable-Baseline3 \cite{stable-baselines3}. A total of 100 environments are simulated in parallel, each with a maximum episode length of 1200 time steps (equivalent to 12 seconds of simulated time). The training ends upon reaching a training budget of 100 million time steps. The hyperparameters used during training is shown in Table \ref{tab:ppo_hyperparameters}. The implementation can be found on GitHub\footnote{\url{https://github.com/tudelft/optimal_quad_control_RL/tree/evolve_drone}}.

\begin{table}[t]
    \centering
    \caption{Drone parameters for domain randomization}
    \begin{tabular}{|c|c|c||c|c|c|}
         \hline
         \textbf{Params} & \textbf{Std} & \textbf{Evo} & \textbf{Params} & \textbf{Std} & \textbf{Evo} \\
         \hline
         $k_{\omega x}$ (10\textsuperscript{-6}) & 0.0 & 2.56 & $k_x$ (10\textsuperscript{-5})& -3.53 & -7.37\\ 
         $k_{\omega y}$ (10\textsuperscript{-6}) & 0.0 & 4.88 & $k_y$ (10\textsuperscript{-5})& -4.99 & -8.55\\ 
         $k_{\omega z}$ (10\textsuperscript{-6}) & 2.55 & 2.47 & $k_z$ (10\textsuperscript{-4})& 0.0 & -2.02\\ 
         \hline
         $k_{p1}$ (10\textsuperscript{-5})& -6.51 & -7.38 & $k_{dp1}$ (10\textsuperscript{-4})& 0.0 & 1.38\\
         $k_{p2}$ (10\textsuperscript{-5})& -6.18 & -4.93 & $k_{dp2}$ (10\textsuperscript{-5})& 0.0 & 9.13\\
         $k_{p3}$ (10\textsuperscript{-5})& 6.21 & 5.26 & $k_{dp3}$ (10\textsuperscript{-4})& 0.0 & 1.15\\
         $k_{p4}$ (10\textsuperscript{-5})& 6.13 & 6.55 & $k_{dp4}$ (10\textsuperscript{-4})& 0.0 & -1.78\\
         \hline
         $k_{q1}$ (10\textsuperscript{-5})& -4.80 & -4.30 & $k_{dq1}$ (10\textsuperscript{-4})& 0.0 & 2.65\\
         $k_{q2}$ (10\textsuperscript{-5})& 5.78 & 5.72 & $k_{dq2}$ (10\textsuperscript{-4})& 0.0 & -4.55\\
         $k_{q3}$ (10\textsuperscript{-5})& -5.24 & -5.89 & $k_{dq3}$ (10\textsuperscript{-4})& 0.0 & 3.02\\
         $k_{q4}$ (10\textsuperscript{-5})& 5.11 & 3.77 & $k_{dq4}$ (10\textsuperscript{-4})& 0.0 & -3.64\\
         \hline
         $k_{r1}$ (10\textsuperscript{-5})& -0.495 & -1.06 & $k_{dr1}$ (10\textsuperscript{-3})& -1.73 & -2.18\\
         $k_{r2}$ (10\textsuperscript{-5})& 0.495 & 1.66 & $k_{dr2}$ (10\textsuperscript{-3})& 1.73 & 2.17\\
         $k_{r3}$ (10\textsuperscript{-5})& 0.495 & 0.776 & $k_{dr3}$ (10\textsuperscript{-3})& 1.73 & 1.91\\
         $k_{r4}$ (10\textsuperscript{-5})& -0.495 & -1.72 & $k_{dr4}$ (10\textsuperscript{-3})& -1.73 & -2.16\\
         \hline
         $\omega_{min}$ (rad/s) & 248.6 & 217.4 & $k_t$ & 0.94 & 0.90 \\
         $\omega_{max}$ (rad/s) & 3276.6 & 3115.2 & $\tau$ & 0.04 & 0.03 \\
         \hline
    \end{tabular}
    \label{tab:drone_parameters}
\end{table}

\begin{table}[t]
    \centering
    \caption{Hyperparameters used for PPO training.}
    \begin{tabular}{c|c}
        \hline
        \textbf{Parameter} & \textbf{Value} \\
        \hline
        Number of parallel environments & 100  \\
        Maximum training steps & 1 $\times$ $10^8$ \\
        Batch size & 5000 \\
        Discount factor ($\gamma$) & 0.999 \\
        Policy network size & [64, 64, 64] \\
        Value network size & [64, 64, 64] \\
        \hline
    \end{tabular}
    \label{tab:ppo_hyperparameters}
\end{table}

\section{Experimental Setup}
\label{sec:experimental_setup}
For the real-world test, we select the most maneuverable evolved drone (Figure \ref{fig:ctrl-drone}) and compare its performance against the standard 220mm wheelbase quadcopter for the drone racing task. A wheelbase of 220mm is selected as it is commonly used in 5-inch racing drones. In order to ensure that the performance differences are primarily due to morphological differences, both the standard and evolved drones utilize identical hardware and the same design philosophy.

Although off-the-shelf 220mm quadcopter frames are readily available, we choose to design and build the frames for both the standard and evolved drones from scratch, in order to ensure that both drones use the same design to maintain fairness in comparison. Figure \ref{fig:intro} shows the constructed drone used in this experiment. The main body is constructed out of 2mm-thick carbon fiber plates, with arm mounting holes specifically milled for each arm angle ($\theta$). Each arm is made of carbon fiber tubes and is cut according to its respective arm length ($l$). To attach the arm to the main body, 3D-printed PLA clamps are used together with aluminum stand-offs, and secured with M3 screws. At the end of each arm, a custom 3D-printed motor mount is designed to set the motors according to their required propeller inclination ($\varphi$) and azimuth ($\psi$). This method of construction ensures a consistent design and modularity across different drone morphologies, resulting in fairer performance comparisons between different configurations.

Both drones are equipped with 2400KV motors with three-bladed 5-inch propellers. For the flight controller, both drones use the STM32F722 microcontroller running INDIflight\footnote{\url{https://github.com/tudelft/indiflight}}, which is a fork of Betaflight. For position and attitude observations, an onboard extended Kalman filter (EKF) fuses IMU readings with measurements from an external motion capture system, more specifically, OptiTrack. The resulting state estimates are then used by the neural network policy. Both drones are also equipped with a Raspberry Pi Zero, which is only used to relay OptiTrack measurements from a ground station to the microcontroller. No computation is performed on the Raspberry Pi Zero or ground station; the neural network policy runs entirely on the microcontroller.

\section{Experimental Results}
\label{sec:experimental_results}
\begin{figure}
    \centering
    \includegraphics[width=0.60\textwidth]{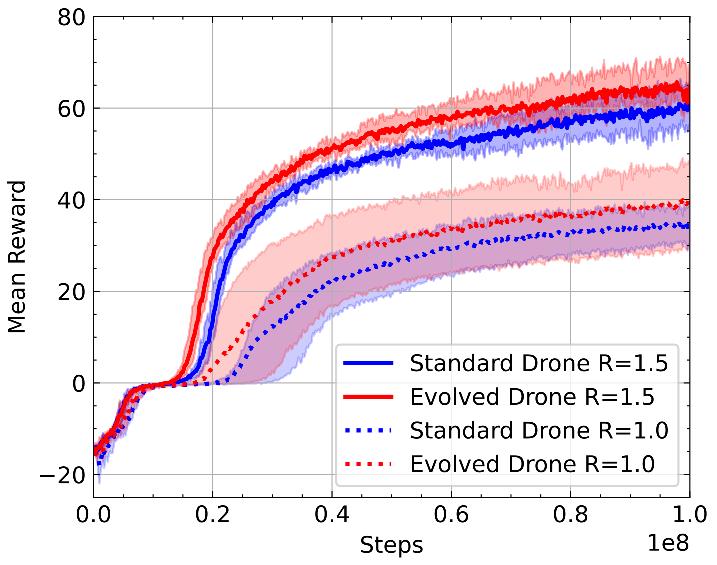}
    \caption{Mean episodic reward over three runs for the standard and evolved drones navigating the R=1.5m and R=1.0m tracks.}
    \label{fig:mean_reward}
\end{figure}

Figure \ref{fig:mean_reward} presents the mean episodic rewards for both drones throughout the training process. As shown in the figure, the evolved drone achieved a higher mean episodic reward than the standard drone at the end of training on both the R=1.5m and R=1.0m. This indicates that the evolved drone is capable of flying faster and covering a greater track distance than the standard drone. Generally, the mean episodic rewards for the R=1.0m track are lower than the rewards for the R=1.5m track. The reduced track size results in tighter turns, forcing the drones to fly at a slower speed in order to stay on the course, resulting in lower track distance and progress reward.

\begin{figure}
    \centering
    \begin{subfigure}[b]{0.60\textwidth}
        \centering
        \includegraphics[width=\textwidth]{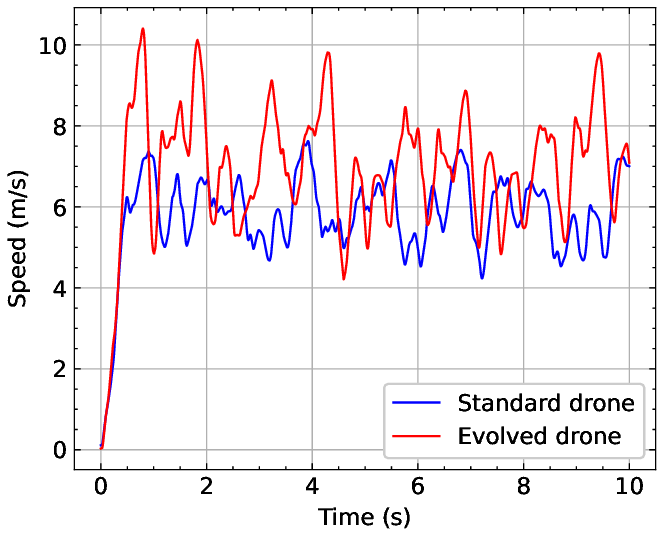}
        \caption{R=1.5 m track.}
        \label{fig:r150_speed}
    \end{subfigure}
    \vfill
    \begin{subfigure}[b]{0.60\textwidth}
        \centering
        \includegraphics[width=\textwidth]{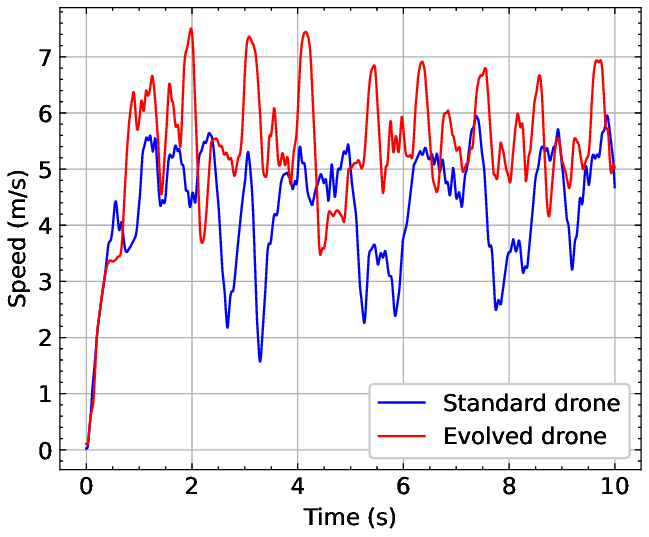}
        \caption{R=1.0 m track.}
        \label{fig:r100_speed}
    \end{subfigure}
    \caption{Speed of the standard (blue) and evolved (red) drones for (a) R=1.5 m track and (b) R=1.0 m track.}
    \label{fig:drone_speed}
\end{figure}

Figures \ref{fig:r150_speed} and \ref{fig:r100_speed} show the speed profiles of both drones from the experiment as they fly along the R=1.5m and R=1.0m tracks, respectively. In general, the evolved drone reaches higher peak velocities compared to the standard drone. Additionally, when the velocity dips during turns, the evolved drone generally maintains higher minimum speeds. This can be attributed to the higher maneuverability, which allows the drone to generate higher moments. Hence, it is able to navigate the turns at a higher speed without drifting off track. As the track size reduces from R=1.5m to R=1.0m, the performance gap between the standard and evolved drone widens, with the evolved drone showing an improvement in average speed of 16.7\% on the larger track and 19.5\% on the smaller track.


\begin{figure}[t]
    \centering
    \includegraphics[width=\linewidth]{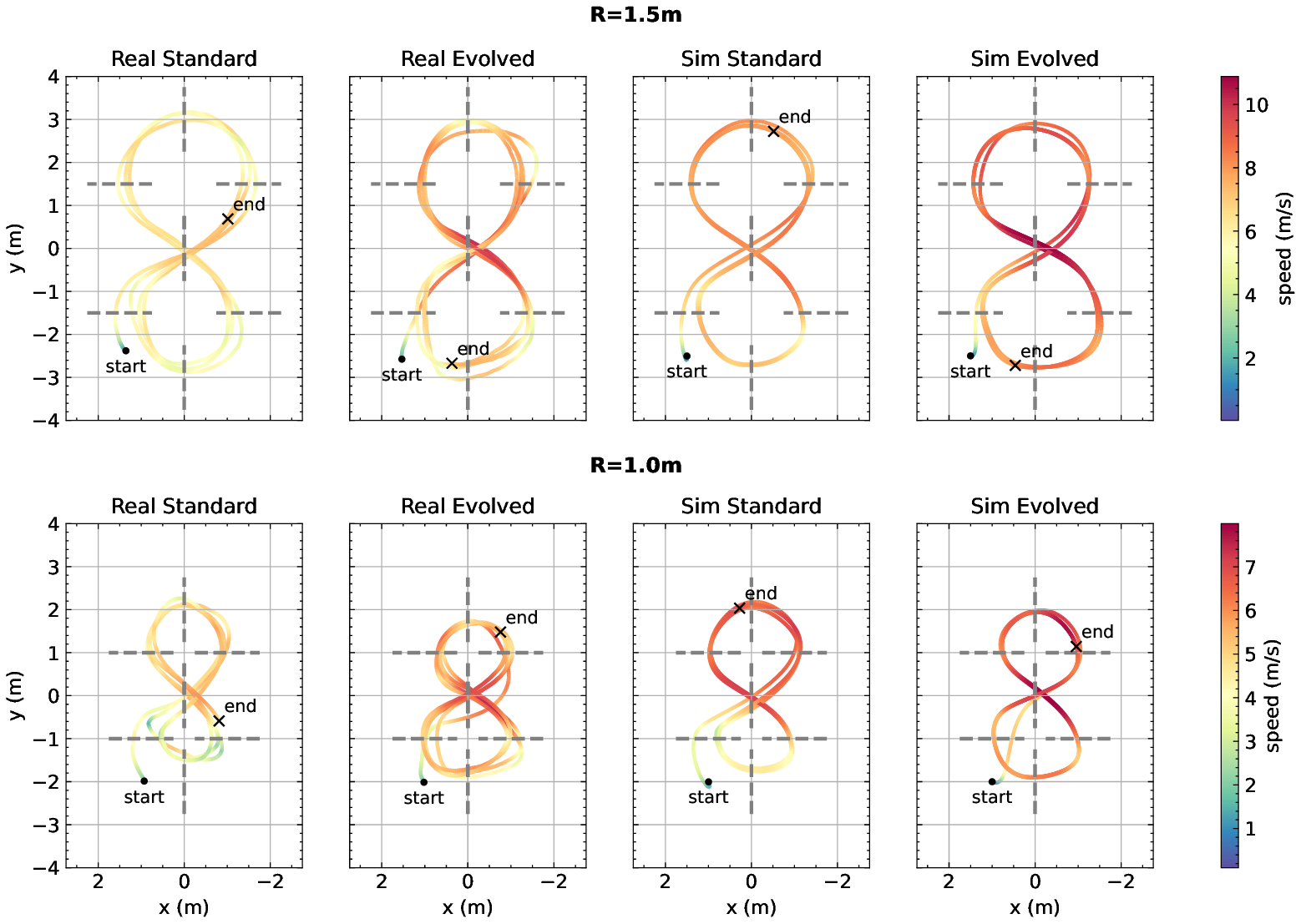}
    \caption{Trajectories of the standard quadcopter and the evolved drone for the R=1.5m (top) and R=1.0m (bottom) tracks. The color of the lines represents the speed of the drones.}
    \label{fig:trajectory}
\end{figure}

Figure \ref{fig:trajectory} shows the trajectories taken by the drones in simulation and the real-world tests. In all cases, the drones decelerate when entering a turn to avoid drifting off the track and accelerate as they exit the turn. As a result, the drones experience their highest speeds as they pass through the middle gate. Generally, the evolved drone achieves higher speeds while flying around the track. When comparing simulation and real-world results, the real trajectories generally follows the trajectories observed in simulation. However, the real world flight path exhibits slightly larger variation, likely due to measurement noise and disturbances. Additionally, the speeds achieved in the simulations are observed to be higher than those in the real-world test.

\begin{figure}
    \centering
    \includegraphics[width=0.60\textwidth]{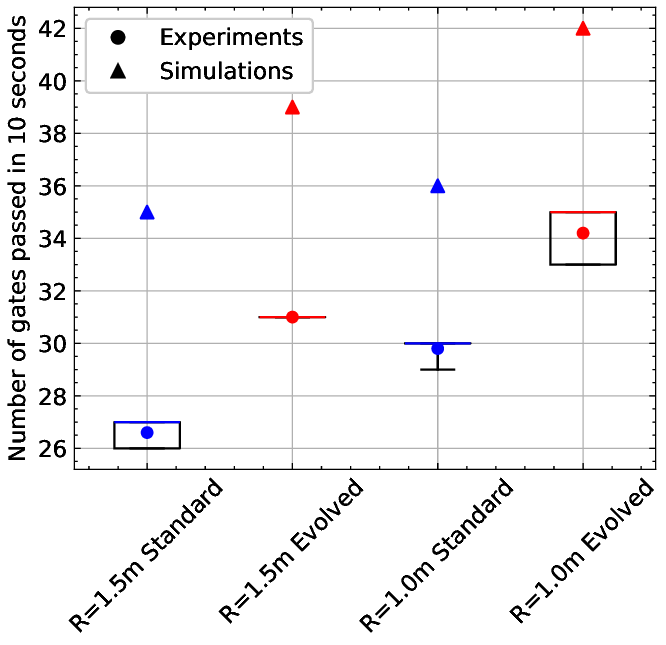}
    \caption{Number of gates passed in 10s by the standard drone (blue) and the evolved drone (red) in both tracks over 5 runs. Experimental results are compared with simulations. Higher numbers are better.}
    \label{fig:gates_passed}
\end{figure}

\begin{figure}
    \centering
    \includegraphics[width=0.60\textwidth]{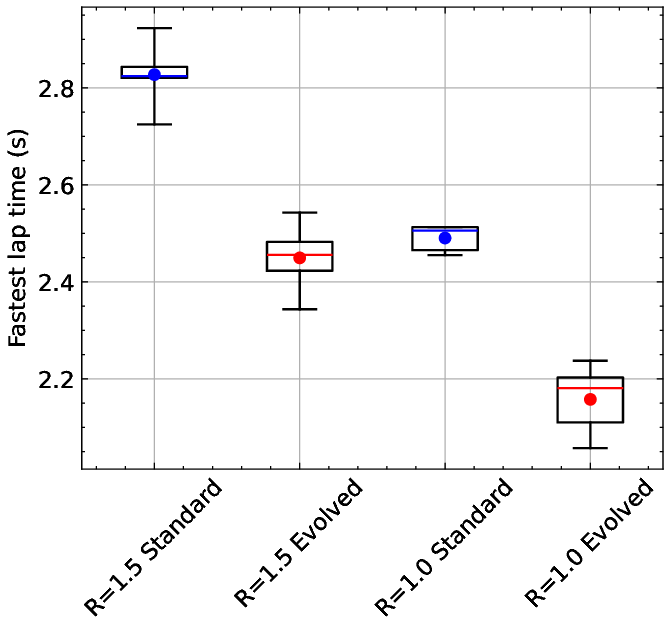}
    \caption{Fastest lap achieved by the standard drone (blue) and the evolved drone (red) in both tracks over 5 runs. Lower times are better.}
    \label{fig:fastest_lap}
\end{figure}

Figure \ref{fig:gates_passed} shows the number of gates passed by each drone in 10 seconds on both tracks. Each run is repeated five times to ensure repeatability. In all cases, the evolved drone managed to pass more gates than the standard quadcopter. As previously mentioned, the simulation results show better performance than real-world results due to noise, disturbances, and modelling errors. Interestingly, although the drones fly slower in the smaller track, they managed to pass more gates compared to the larger track, primarily due to the reduced gate-to-gate distance on the smaller track. Figure \ref{fig:fastest_lap} shows the fastest lap time achieved by each drone for both tracks, with each run being repeated five times. Similarly, in both tracks, the evolved drone outperforms the standard quadcopter, achieving faster lap times. The evolved drone demonstrates a 13.4\% improvement in average fastest lap time on both tracks.

\section{Conclusions}
\label{sec:conclusion}
In this paper, multi-objective optimization of drone bodies is performed using the NSGA-II algorithm to maximize the thrust-to-weight ratio and maneuverability while minimizing their size. The NSGA-II algorithm generated a range of drone bodies that can outperform a standard quadcopter in at least one of the three objectives, with the smallest drone being 4.8\% smaller, and the most maneuverable and highest thrust-to-weight ratio being 489.2\% and 23\% better than a standard quadcopter, respectively. The Pareto front provides insights into the performance trade-offs when modifying the morphology of the drones. Generally, drones with more propellers can generate a higher thrust-to-weight ratio at a cost of larger size and lower maneuverability. On the other hand, possessing fewer propellers results in smaller and more maneuverable drones at the cost of a lower thrust-to-weight ratio. At the same time, tilting the propellers generally improves the maneuverability of the drone by increasing control authority in the yaw axis. Thereafter, we take the most maneuverable drone and train reinforcement policies to compare its performance with a standard quadcopter in a real-world drone racing task. The experimental results show that for both the larger R=1.5m and the smaller R=1.0m tracks, the evolved drone outperforms the standard quadcopter in terms of flight speed, number of gates passed and lap times. This shows that evolving morphologies using low-level metrics, such as maneuverability, translate to improved real-world performance.


\section*{Acknowledgments}
This work was supported by the European Commission Horizon project SPEAR under grant agreement 101119774.

\bibliographystyle{elsarticle-num} 
\bibliography{biblio}

\end{document}